\documentclass[conference]{IEEEtran}
\IEEEoverridecommandlockouts
\usepackage{cite}
\usepackage{amsmath,amssymb,amsfonts}
\usepackage{algorithmic}
\usepackage[dvipdfmx]{graphicx}
\usepackage{graphicx}
\usepackage{textcomp}
\usepackage{xcolor}
\usepackage{algorithm} 
\usepackage{subcaption}
\usepackage{ntheorem}
\usepackage{hyperref}
\usepackage{textcomp}
\usepackage{comment}
\usepackage{color}
\usepackage{soul}
\usepackage{url}
\usepackage[misc]{ifsym}

\theoremseparator{:}

\newboolean{showcomments}
\setboolean{showcomments}{true} 
\ifthenelse{\boolean{showcomments}}
{\newcommand{\nb}[2]{
		\fcolorbox{black}{yellow}{\bfseries\sffamily\scriptsize#1}
		{\sf\small$\blacktriangleright$\textit{#2}$\blacktriangleleft$}
	}
	
}
{\newcommand{\nb}[2]{}
	
}

\makeatletter
\newcommand{\figcaption}[1]{\def\@captype{figure}\caption{#1}}
\newcommand{\tblcaption}[1]{\def\@captype{table}\caption{#1}}
\makeatother

\def\BibTeX{{\rm B\kern-.05em{\sc i\kern-.025em b}\kern-.08em
    T\kern-.1667em\lower.7ex\hbox{E}\kern-.125emX}}
\begin{document}

\newcommand\jialong[1]{\nb{Jialong}{#1}}
\newcommand\sherry[1]{\nb{Sherry}{#1}}
\newcommand\kenji[1]{\nb{Kenji}{#1}}

\title{
Towards Context-aware Support for Color Vision Deficiency: An Approach Integrating LLM and AR 
}

\author{
    \IEEEauthorblockN{
        Shogo Morita\textsuperscript{1}, 
        Yan Zhang\textsuperscript{2}, 
        Takuto Yamauchi\textsuperscript{3}, 
        Sinan Chen\textsuperscript{4}, 
        Jialong Li\textsuperscript{3}, 
        Kenji Tei\textsuperscript{1}
    }
    \IEEEauthorblockA{\textit{\textsuperscript{1} School of Computing, Tokyo Institute of Technology, Tokyo, Japan}}
    \IEEEauthorblockA{\textit{\textsuperscript{2}International Research Center for Neurointelligence, the University of Tokyo, Tokyo, Japan}}
    \IEEEauthorblockA{\textit{\textsuperscript{3}Department of Computer Science and Engineering, Waseda University, Tokyo, Japan}}
    \IEEEauthorblockA{\textit{\textsuperscript{4}Center of Mathematical and Data Sciences, Kobe University, Kobe, Japan}}
    \IEEEauthorblockA{Correspondence: lijialong@fuji.waseda.jp}
}

\maketitle

\begin{abstract}
People with color vision deficiency often face challenges in distinguishing colors such as red and green, which can complicate daily tasks and require the use of assistive tools or environmental adjustments. Current support tools mainly focus on presentation-based aids, like the color vision modes found in iPhone accessibility settings. However, offering context-aware support, like indicating the doneness of meat, remains a challenge since task-specific solutions are not cost-effective for all possible scenarios. To address this, our paper proposes an application that provides contextual and autonomous assistance. This application is mainly composed of: (i) an augmented reality interface that efficiently captures context; and (ii) a multi-modal large language model-based reasoner that serves to cognitize the context and then reason about the appropriate support contents.
Preliminary user experiments with two color vision deficient users across five different scenarios have demonstrated the effectiveness and universality of our application.
\end{abstract}


\section{Introduction}
Individuals with color vision deficiency often struggle to distinguish certain colors, particularly red and green. This inability creates various challenges in their daily lives, such as identifying traffic lights, choosing clothing, or engaging in tasks that require color differentiation.  
As a result, they may need to rely on special assistive tools or adjust environmental settings to reduce obstacles in everyday life.

Currently, tools for such groups tend to focus on adjustments in display or presentation, like the VoiceOver function and the option of high contrast settings found in iPhone’s accessibility. Additionally, many studies have explored enhancing readability by recoloring, for instance, \cite{recolor} utilizes Transformer in image recoloring for color vision deficiency compensation.

However, a more complex challenge is supporting individuals with color vision deficiency from a context perspective rather than just a display perspective \cite{SEAMS24_li}. For example, individuals with visual impairments may not be able to determine the doneness of meat on a grill, requiring assistance from shop attendants or companions to inform them ``if the meat is cooked'' \cite{Kishida2024}. While specialized supporting systems have been developed for some frequent scenarios, such as cloth coordination \cite{ClothMatch}, it is not economically feasible to develop dedicated systems for every possible scenario.

In this paper, we introduce a general context-aware support approach for individuals with color vision deficiency.
Specifically, as a user interface, we integrate augmented reality (AR) to capture real-world data and display supportive content. 
Internally, we develop a set of prompt strategies to leverage the expansive understanding and potent reasoning abilities of multi-modal large language models (LLMs) to generate appropriate content. This approach aims to provide more autonomous and context-sensitive support for everyday challenges faced by those with color vision deficiencies.

\section{Proposed Application}
\begin{figure}[thb!]
    \centering
    \includegraphics[width=\linewidth]{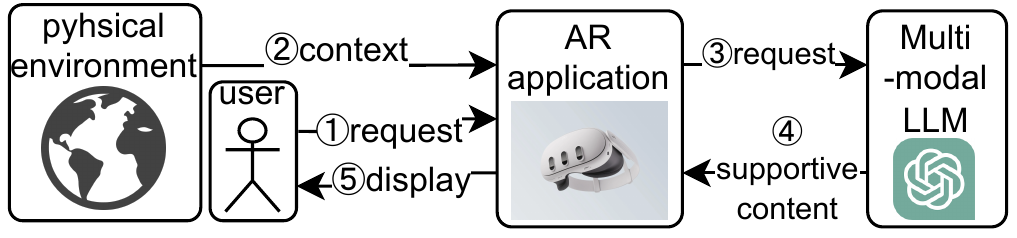}
    \caption{Overview of the Application Interaction.}
    \label{overview}
\end{figure}

\textbf{Application Overview}.
Figure~\ref{overview} illustrates the workflow of our application. Initially, the AR application receives a support request from the user. Subsequently, the application utilizes a camera to collect environmental data. This data is then integrated with other relevant information to create a prompt for invoking a multi-modal LLM. Following this, the LLM generates supportive content, and this content is finally displayed to the user through the AR application.

\textbf{AR Interface}.
The AR interface serves two primary functions: input and output. For input, it utilizes a camera to capture environmental information around the user, supplemented by a microphone or a button to record user requests. We categorize user requests into two types. The first involves direct input from the user specifying the needed support, such as saying "Please tell me the color of the traffic light" through the microphone. The second type is a simplified interaction where the user only presses a button indicating the need for help, with the specifics inferred by the LLMs. This simplified approach aims to reduce the user’s operational burden and enhance privacy. For output, the system displays supportive content on the AR device screen, presented in a textbox at the bottom of the user's field of view. We chose AR over mobile devices like smartphones because AR offers a more intuitive and efficient interaction experience, capturing directly the area the user is focusing on and overlaying digital information in real time. While glasses-type AR devices would be ideal, we used the Meta Quest 3 for its cost-effectiveness and comprehensive development kit.

\textbf{LLM-based support contents generation}.
We employ multi-modal LLMs to generate supportive content. To enhance the effectiveness and accurancy of content generation, we carefully designed prompts. Initially, as context information, we provide basic information about the user and their specific characteristics, such as reduced sensitivity to red light in individuals with Protanomaly. Secondly, we implement a Chain-of-Thought (CoT) prompting approach, including the following four steps: (i) analyzing the current environmental situation; (ii) determining the user's intent and the type of help required; (iii) generating concise supportive content, limited to 10 words to ensure readability; and (iv) identifying key terms for emphasis presentation (e.g., bold text) within the supportive content. The first three steps of the CoT are modeled after the human cognitive decision-making process, which includes perception of the environment, high-level decision-making, and low-level realization.

\section{Preliminary Evaluation}
\begin{figure}[thb!]
    \centering
    \includegraphics[width=0.8\linewidth]{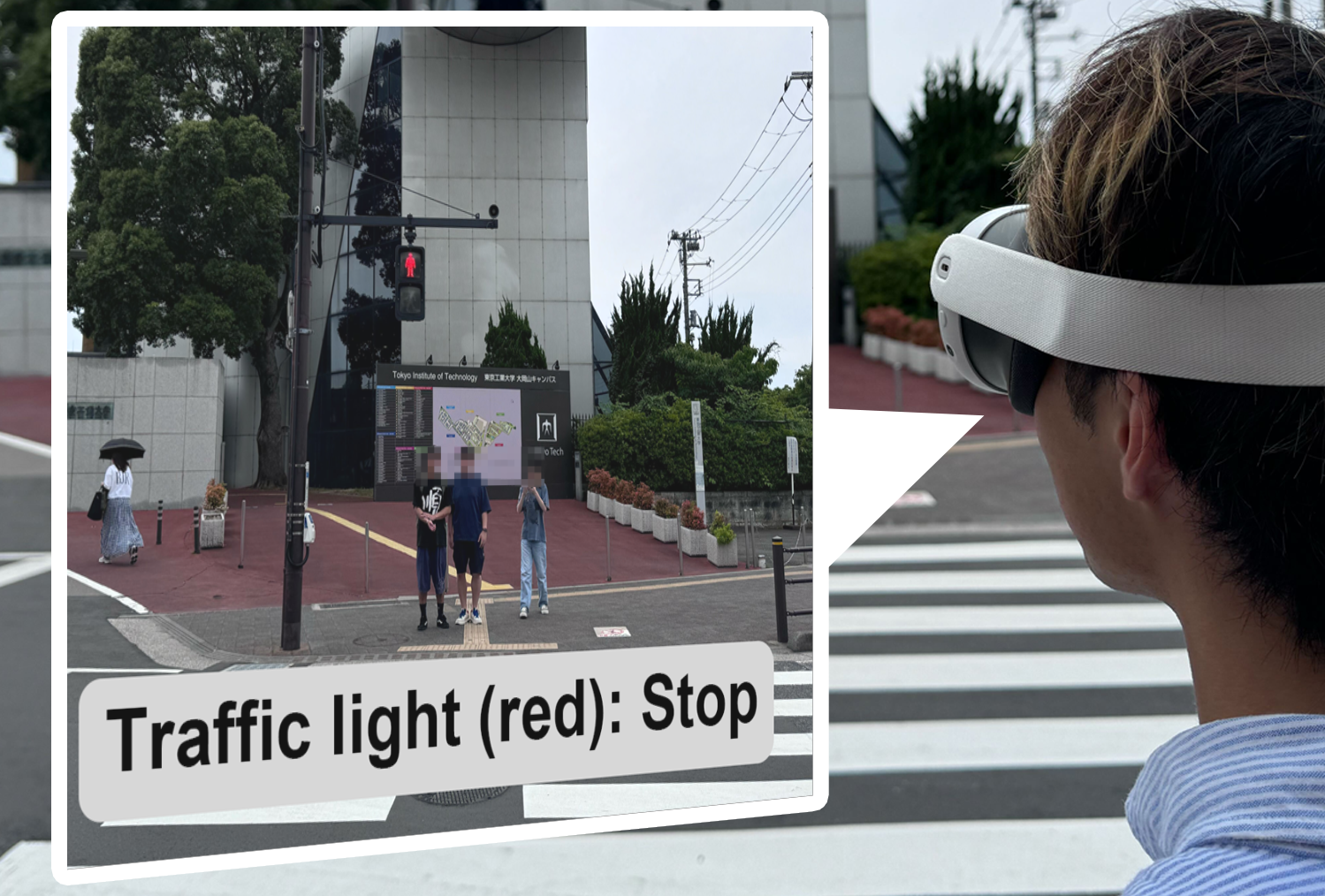}
    \caption{Scene of Usage and Screen in Device (bottom left).}
    \label{example}
\end{figure}

\textbf{Experiment setting}. 
To evaluate the effectiveness of our application, we conducted an experiment with two actual color vision deficient participants from a university campus. Informed consent forms were distributed before participation, and data handling complied with the ethics board recommendations of the university. Specifically, we had the participant utilize our application, followed by an in-depth interview to gather qualitative and practical feedback.
To validate the universality of our application, affirming its applicability across various everyday scenarios, we employed five distinct experimental scenarios. These scenarios were: (i) identifying traffic lights; (ii) judging the doneness of meat; (iii) selecting ripe fruits; (iv) coordinating clothing; and (v) reading color-coded signs in public transportation. For each scenario, we conducted tests in two different environments.
For multi-modal LLM, we selected the most advanced and popular GPT4-o.

\textbf{Experiment results}. 
Firstly, for the objective accuracy, in ten tests (five scenarios times two different environments), the multi-modal LLMs were able to correctly recognize the environment context and user intentions, and provide accurate supportive contents.
Secondly, in actual user experience and user interviews, users acknowledged the practical effectiveness of our application (8.5 out of 10 in average). This includes (i) the convenience provided by the AR interface, which is more straightforward than having to align and photograph a target with a smartphone; and (ii) the accuracy of the LLMs in inferring user intentions, which significantly reduces the user's effort and increases response efficiency as users do not need to provide specific vocal instructions.

\textbf{Discussion and Limitations}.
We believe the preliminary user experiments have already demonstrated the effectiveness and usability of our application. However, several limitations were identified. Firstly, participants noted that the disadvantages for color vision deficient individuals are more pronounced in the workplace  than in everyday life, suggesting that further evaluation of the application's usefulness in working settings is necessary. Secondly, the risk associated with result accuracy was highlighted, as LLMs cannot guarantee the accuracy of their outputs. However, participants also indicated that they often integrate information from other sources (such as observing the behavior of others at a traffic light) to make judgments. Thus, the supportive context generated by our application, serving as one of the reference sources, is still beneficial. Thirdly, in complex scenarios, such as identifying the doneness of multiple pieces of meat on a baking tray, there is a need for a more precise and intuitive explanation. This suggests further enhancement of the content visualization, such as marking numbers on each piece of meat.

\section{Conclusion and Future Work}
\label{sec: conclusion}
This paper introduces a general context-aware support approach for individuals with color vision deficiency, employing AR and multi-modal LLMs. Preliminary user experiments involving two color vision deficient users across five everyday scenarios demonstrate the universality and effectiveness of our approach.  
Future research will focus on two main areas. First, we plan to improve result accuracy through appropriate prompt engineering.
Second, we aim to conduct further experiments in a broader range of everyday and workplace scenarios with a larger number of participants to more comprehensively validate our approach.

\bibliographystyle{IEEEtran}
\bibliography{bib}

\begin{thebibliography}{1}
\providecommand{\url}[1]{#1}
\csname url@samestyle\endcsname
\providecommand{\newblock}{\relax}
\providecommand{\bibinfo}[2]{#2}
\providecommand{\BIBentrySTDinterwordspacing}{\spaceskip=0pt\relax}
\providecommand{\BIBentryALTinterwordstretchfactor}{4}
\providecommand{\BIBentryALTinterwordspacing}{\spaceskip=\fontdimen2\font plus
\BIBentryALTinterwordstretchfactor\fontdimen3\font minus \fontdimen4\font\relax}
\providecommand{\BIBforeignlanguage}[2]{{%
\expandafter\ifx\csname l@#1\endcsname\relax
\typeout{** WARNING: IEEEtran.bst: No hyphenation pattern has been}%
\typeout{** loaded for the language `#1'. Using the pattern for}%
\typeout{** the default language instead.}%
\else
\language=\csname l@#1\endcsname
\fi
#2}}
\providecommand{\BIBdecl}{\relax}
\BIBdecl

\bibitem{recolor}
L.~Chen, Z.~Zhu, W.~Huang, K.~Go, X.~Chen, and X.~Mao, ``Image recoloring for color vision deficiency compensation using swin transformer,'' \emph{Neural Comput. Appl.}, 2024.

\bibitem{SEAMS24_li}
J.~Li, M.~Zhang, N.~Li, D.~Weyns, Z.~Jin, and K.~Tei, ``Exploring the potential of large language models in self-adaptive systems,'' \emph{SEAMS}, 2024.

\bibitem{Kishida2024}
N.~Kishida, ``Today, whose perspective? - color blindness and me,'' \url{https://heart-design.jp/column/%E4%BB%8A%E6%97%A5%E3%81%AF%E8%AA%B0%E7%9B%AE%E7%B7%9A%EF%BC%9F/}, 2024, accessed: 2024-06-20.

\bibitem{ClothMatch}
Y.~Tian and S.~Yuan, ``Clothes matching for blind and color blind people,'' \emph{Computers Helping People with Special Needs}, 2010.

\end{thebibliography}
\end{document}